\pdfoutput=1

\documentclass[11pt]{article}

\usepackage[]{acl}

\usepackage{times}
\usepackage{latexsym}
\usepackage{diagbox}

\usepackage[T1]{fontenc}

\usepackage[utf8]{inputenc}

\usepackage{microtype}

\usepackage{inconsolata}

\usepackage{times}
\usepackage{latexsym}

\usepackage[T1]{fontenc}

\usepackage[utf8]{inputenc}

\usepackage{microtype}
\usepackage{times}
\usepackage{url}
\usepackage{latexsym}

\usepackage{graphicx}
\graphicspath{ {./} }
\usepackage[table,xcdraw]{}
\usepackage{microtype}
\usepackage{booktabs}
\usepackage{subfigure}
\usepackage{graphics}
\usepackage{amsmath}
\usepackage{color}
\usepackage{diagbox}
\usepackage{amssymb}
\usepackage{multirow} 
\usepackage{makecell}
\usepackage{tabularx}
\usepackage{threeparttable}
\usepackage{amssymb}
\usepackage{marvosym}
\usepackage{subfigure}
\usepackage{soul}
\usepackage{times}
\usepackage{latexsym}
\usepackage{xcolor}

\usepackage{inconsolata}
\title{Multi-Label Classification for Implicit Discourse Relation Recognition}

\author{Wanqiu Long\textsuperscript{1} \and N. Siddharth\textsuperscript{1,2} \and Bonnie Webber\textsuperscript{1} \\
  \textsuperscript{1}University of Edinburgh, Edinburgh, UK \\
  \textsuperscript{2}The Alan Turing Institute \\
  \texttt{Wanqiu.long@ed.ac.uk, n.siddharth@ed.ac.uk, Webber.Bonnie@ed.ac.uk}
}

\begin{document}

\maketitle

\begin{abstract}
Discourse relations play a pivotal role in establishing coherence within textual content, uniting sentences and clauses into a cohesive narrative. The Penn Discourse Treebank (PDTB) stands as one of the most extensively utilized datasets in this domain. In PDTB-3 \cite{webber2019penn}, the annotators can assign multiple labels to an example, when they believe that multiple relations are present. Prior research in discourse relation recognition has treated these instances as separate examples during training, and only one example needs to have its label predicted correctly for the instance to be judged as correct. However, this approach is inadequate, as it fails to account for the interdependence of labels in real-world contexts and to distinguish between cases where only one sense relation holds and cases where multiple relations hold simultaneously. In our work, we address this challenge by exploring various multi-label classification frameworks to handle implicit discourse relation recognition. We show that multi-label classification methods don’t depress performance for single-label prediction. Additionally, we give comprehensive analysis of results and data. Our work contributes to advancing the understanding and application of discourse relations and provide a foundation for the future study.
\end{abstract}

\section{Introduction}
Discourse relations serve as a fundamental framework for creating coherence among sentences and clauses within a text. The automated identification of the discourse relations, which connect sentences and clauses, holds significant importance for a wide range of downstream Natural Language Processing (NLP) tasks including text summarization \cite{huang-kurohashi-2021-extractive}, question answering \cite{pyatkin-etal-2020-qadiscourse}, and event relation extraction \cite{tang2021discourse}. Penn Discourse Treebank (PDTB) \cite{webber2019penn,prasad-etal-2008-penn} is one of the most widely used datasets for this task. It includes over 2000 documents from the
\textit{Wall Street Journal} published in 1989, which have been manually annotated with discourse relations.



With regard to PDTB annotation, annotators can assign multiple sense labels to an example when they believe that all hold simultaneously. Consider the following example from PDTB-3.
 
 \begin{enumerate}
 
\item[(1)]  [In the past decade, Japanese manufacturers concentrated on domestic production for export]. [In the 1990s, spurred by rising labor costs and the strong yen, these companies will increasingly turn themselves into multinationals with plants around the world]. (wsj\_0043).
\textbf{Labels:} Asynchronous and Concession

\end{enumerate}

This example demonstrates the simultaneous presence of two distinct discourse relations. The annotator wanted to capture the sense that the first sentence described an earlier time period and the second, a subsequent time period. But the annotator also wanted to capture the sense that DESPITE Japan's concentration on domestic production for export in the earlier time period, the future will be different -- i.e. conceding what was true in the 1980s, and asserting that the future will be different. Therefore, the pair of arguments in the example is linked by two discourse relations that capture different dimensions of the relationship between them. Understanding these complex relations is crucial for the interpretation of discourse.






However, for those instances with two annotated labels, all previous work on discourse relation recognition treat them as separate and different examples during training, and at test time, a prediction matching one of the gold types is taken as the correct answer. 

Nevertheless, by treating instances with multiple labels as separate examples, the model may not effectively capture the inherent complexity of discourse relations. Real-world texts often contain multiple layers of meaning, and forcing the model to treat them as distinct instances may oversimplify the problem. This approach may not reflect the true nature of discourse relations, where multiple relations can coexist between two arguments. Moreover, when treating instances with multiple labels as separate examples, the model loses valuable contextual information about how these relations interact and influence each other. Furthermore, treating multi-label instances as separate examples might introduce training ambiguity and lead to conflicting patterns, impairing the model's ability to generalize to new examples. Besides, when implementing discourse relations recognition in downstream tasks, the inability to recognize multi-labels could potentially lead to adverse effects. For Example 1, if the model fails to identify both Concession and Asynchronous relation simultaneously, it may struggle to respond to questions concerning temporal order and contingency relation concurrently.



To address these negative impacts, we explored multi-label classification as a more effective way to capture the complexity of discourse relations. To date, implicit discourse relation recognition has not been approached as a multi-label classification task, despite advancements facilitated by pretrained language models \cite{zhang-etal-2021-context,long-webber-2022-facilitating, wu-etal-2023-connective}. The current study compares three multi-label classification methods for IDRR, alongside conducting an in-depth analysis of the results and data. The main contributions of our work are as follows:
\begin{itemize}
\setlength{\itemsep}{0pt}
\setlength{\parsep}{0pt}
\setlength{\parskip}{0pt}

    \item  This is the first study to treat implicit discourse relation recognition as a multi-label classification issue, marking a novel approach in addressing the complexity of discourse relations more effectively compared to traditional single-label prediction methods.    
    \item We explore different multi-label classification methods for IDRR, and we show that a multi-label classifier can even demonstrate better performance than a classifier trained on examples in which multiple labels were split into two distinct and unrelated examples.
    \item We provide a fine-grained analysis of results and conduct some methodological exploration, which can give valuable insights and pave the way for the future study.

\end{itemize}

\section{Related Work}
\textbf{Multi-label Classification} Multi-label classification has been widely adopted across different NLP applications, including intent detection \cite{Ouyang2022TowardsMU}, emotion classification \cite{yu-etal-2018-improving,alhuzali-ananiadou-2021-spanemo}, and document classification \cite{xiao-etal-2019-label,liu-etal-2022-bert,lin-etal-2023-effective}. Despite the potential for richer text understanding, multi-label classification has been unexplored in discourse relation recognition. In this work, we expand the use of multi-label classification to implicit discourse relation recognition to enhance the accuracy and depth of NLP applications from richer text understanding.

\noindent
\textbf{Multi-label Study on Discourse Relation Recognition} Two notable studies on multiple label examples for discourse relation recognition are: 1) \newcite{marchal-etal-2022-establishing}, which explores inter-coder agreement in multiple label examples, revealing instances where annotators assign more than one relation to a single example from various genres. 2) \newcite{scholman-etal-2022-discogem} introduces a crowdsourced corpus containing 6,505 implicit discourse relations across three genres. Annotators in this corpus assign a single discourse sense based on their perspective, resulting in varying interpretations. Nevertheless, the identification of implicit discourse relations has not been addressed in the context of a multi-label classification framework.



\section{Dataset and Evaluations}
We employed PDTB-3 \cite{webber2019penn} for our evaluation. PDTB-3 represents an advancement over PDTB-2 \cite{prasad-etal-2008-penn}, offering a more extensive collection of annotated multi-label examples. In our study, we concentrated exclusively on implicit discourse relations, disregarding those with explicit connectives. And we consider both intra-sentential  and inter-sentential implicits.

About 5\% of PDTB-3 implicit discourse relations receive multiple labels, which corresponds to instances with two annotated labels. We treated such instances as single examples with multi-labels during training, and during testing, predictions were considered correct only if they match the specific label.

While previous studies \cite{ji-eisenstein-2015-one, bai-zhao-2018-deep, xiang-etal-2022-connprompt} typically allocate Sections 2-20 of PDTB for training, Sections 0-1 for validation, and Sections 21-22 for testing, the limited size of the test set poses challenges, particularly for rare label and label pairs within the dataset. Acknowledging the concerns raised by \citeauthor{shi-demberg-2017-need} regarding label sparsity, we addressed this issue by employing cross-validation for Level-2 classification. In line with the methodology proposed by \citeauthor{kim-etal-2020-implicit}, we adopted a cross-validation approach at the section level. We divided PDTB-3 into 12 folds, with each fold partitioned into 21 sections for training, two for development, and two for testing. By splitting the data at the section level, we can preserve the inherent paragraph and document structures, ensuring that data from the same sections are grouped together in the same pool.



Following the work in multi-label classification for other tasks \cite{Tsai2019OrderfreeLA,Zhang2021EnhancingLC}, we adopt F1 scores \cite{Manning2008IntroductionTI} as our main evaluation metric. Precision, Recall and Hamming loss \cite{Schapire1998ImprovedBA} are reported in the Appendix. We present the macro-averaged results of F1 scores, Precision and Recall.

\section{Performance Comparison: Evaluating the Effectiveness of Different Methods}


\subsection{Methods}
Our work has explored three different multi-label classification techniques, two encoder-only methods and one encoder-decoder method. 
 
\subsubsection{Method 1} 






The output vector corresponding to the [CLS] token aggregates input features and is used for classification. We employ RoBERTa for text representation learning, and add a classification head  $W_c \in \mathbb{R}^{H \times |C|}$ on top of the [CLS] token to do classiciation. $H$ is the dimension size of [CLS] representation and $C$ represents the number of classes. We use $y \in \mathbb{R}^{|C|}$ to denote the ground-truth label for an example, where $y \in \{0, 1\}^{|C|}$. 



The model is trained using sigmoid binary cross-entropy loss. If the predicted probability of a label surpasses 0.5, it is regarded as a predicted label.


\subsubsection{Method 2} 
This method resembles Method 1, with several key distinctions. Rather than employing a single classification head to handle all labels, we utilize multiple classification heads $W_{c_i} \in \mathbb{R}^{H \times 2}$, each $c_i$ dedicated to the $i$-th specific label and treating them as individual binary classification tasks. In contrast to Method 1, which utilizes sigmoid binary cross-entropy loss, we employ softmax cross-entropy for loss calculation here. The loss for each label is computed independently, and subsequently, the mean of these individual losses is used to update the model. If the predicted probability of a label is greater than 0.5, it is considered a predicted label.

\subsubsection{Method 3} 
In this approach, we use a sequence-generating model that processes input text token by token, predicting labels sequentially while considering previously predicted labels. Our method is similar to the one described in \citet{inbook}. We utilize RoBERTa's last transformer block to generate word vectors and use RoBERTa's [CLS] token embedding as the initial hidden state for our decoder, which is a Gated Recurrent Unit (GRU) in our case. Our model also incorporates a dot-product attention mechanism between encoder and decoder. We train the final model to minimize the cross-entropy objective loss for a given $x$ and ground-truth labels $\{t_1^*, t_2^*, \ldots, t_k^*\} \in \mathcal{L}$:

\[
\mathcal{L}_{CE}(\theta) = -\sum_{i=1}^{k} \log P_\theta(t_i^* | x, t_{1:i-1}^*)
\]

During inference, we conduct a beam search to identify candidate sequences with the lowest objective scores among the paths that conclude with the <eos> token. The beam size is set to 4.

\begin{table*}[ht]
\footnotesize
\centering
\label{tab:comparison}
\begin{tabular}{lccc}
\toprule
Label & Method 1& Method 2 & Method 3 \\
\midrule
Concession & 50.98 $\pm$ 5.06 & 51.59 $\pm$ 4.61 & 50.86 $\pm$ 2.91 \\
Contrast & 50.58 $\pm$ 3.40 & 50.82 $\pm$ 2.99 & 48.49 $\pm$ 3.46 \\
Cause & 65.57 $\pm$ 1.76 & 65.15 $\pm$ 1.74 & 65.34 $\pm$ 2.19 \\
Cause+Belief & 0.00 $\pm$ 0.00 & 0.00 $\pm$ 0.00 & 3.04 $\pm$ 5.58 \\
Condition & 75.99 $\pm$ 5.95 & 80.97 $\pm$ 7.51 & 78.01 $\pm$ 10.00 \\
Purpose & 92.50 $\pm$ 2.01 & 92.68 $\pm$ 2.34 & 92.58 $\pm$ 2.21 \\
Conjunction & 62.12 $\pm$ 3.05 & 63.32 $\pm$ 3.34 & 62.11 $\pm$ 3.08 \\
Equivalence & 12.99 $\pm$ 7.56 & 14.55 $\pm$ 8.43 & 17.85 $\pm$ 6.42 \\
Instantiation & 58.76 $\pm$ 4.02 & 59.67 $\pm$ 5.73 & 58.49 $\pm$ 6.15 \\
Level-of-detail & 50.93 $\pm$ 3.97 & 51.80 $\pm$ 3.80 & 51.09 $\pm$ 2.26 \\
Manner & 58.76 $\pm$ 11.39 & 58.60 $\pm$ 13.47 & 23.23 $\pm$ 7.94 \\
Substitution & 64.11 $\pm$ 10.35 & 62.35 $\pm$ 7.83 & 54.46 $\pm$ 9.18 \\
Asynchronous & 62.46 $\pm$ 4.01 & 62.10 $\pm$ 3.88 & 61.20 $\pm$ 3.72 \\
Synchronous & 27.40 $\pm$ 9.48 & 30.28 $\pm$ 7.30 & 30.30 $\pm$ 6.96 \\
\hline
Total & 52.37 $\pm$ 1.62 & 53.13 $\pm$ 0.92 & 49.79 $\pm$ 1.12 \\
\bottomrule
\end{tabular}
\caption{A Comparison of Macro-F1 scores across different methods by using RoBERTa$_{\text{base}}$. We use cross-validation at section level for the Level-2 classification. The standard deviations across 12 folds are reported. ``Total'' here refers to the average scores for all labels.}
\end{table*}

\begin{table*}[htbp]
\footnotesize
\centering
\begin{tabular}{l|c|c|c|c|c|c}
\hline
& \multicolumn{2}{c|}{\textbf{Method 1}} & \multicolumn{2}{c|}{\textbf{Method 2}} & \multicolumn{2}{c}{\textbf{Method 3}} \\
\hline
\diagbox{Num. prediction}{Num. gold} & \textbf{2 } & \textbf{1 } & \textbf{2 } & \textbf{1 } & \textbf{2 } & \textbf{1 } \\
\hline
\multicolumn{1}{c|}{2} & 395 & 806 & 405 & 983 & 379 & 585 \\
\hline
\multicolumn{1}{c|}{1} & 506 & 17780 & 498 & 17614 & 563 & 19396 \\
\hline
\multicolumn{1}{c|}{0} & 41 & 1395 & 39 & 1383 & 0 & 0 \\
\hline
\end{tabular}
\caption{Comparative analysis of predicted label counts for instances with one and two gold labels across Method 1, Method 2, and Method 3. `Num. Prediction''  denotes the number of labels predicted by each method, while ``Num. Gold'' represents the number of gold-standard labels.}
\label{tab:label_prediction_comparison}
\end{table*}



\subsubsection{Implementation Details}
For all experiments across all methods, we use RoBERTa as the pretrained language model. The max sequence length is set to 512. For all experiments, we adopt Adam \cite{kingma2017adam} with the learning rate of $1e{-}5$ and the batch size of 64 to update the model. The maximum training epoch is set to 20 and the wait patience for early stopping is set to 10. All experiments are performed with 1× 80GB NVIDIA A100 GPU.

\subsection{Results}
\subsubsection{Performance for Each label}
Table 1 displays F1 scores for each level-2 label across three methods at section-level cross-validation. Method 2 outperforms the others. Labels like ``Cause'', ``Condition'', ``Purpose'', and ``Conjunction'' consistently perform well across all methods. However, ``Cause+Belief'', ``Synchronous'' and ``Equivalence'' consistently receive lower F1 score, indicating they are more challenging. Method 3 lags on ``Manner'' and ``Substitution'', often missing ``Manner'' in multi-label instances and falsely predicting ``Substitution''.



\subsubsection{Count of Predicted Labels}
Table 2 displays the distribution of the number of the predicted labels for examples with one or two gold labels across the three methods. We did not impose a limit on the number of predicted labels. However, none of the examples received more than two labels for any method, likely due to the data not containing examples with more than two labels. Analyzing the table, we find that \textbf{distinguishing one or two labels is challenging}, as over half of multi-label examples receive only one label, while more than 5\% of single-label examples get two labels. Method 2 tends to assign two labels more often. Method 3 consistently assigns at least one label to all examples, aligning with PDTB's typical one or two annotated labels per example.



\subsubsection{Predictions for multi-label examples}
Table 3 evaluates the performance of the three methods on multi-label examples, namely those examples with two labels. The methods can get: both labels correct, one label correct, both labels incorrect, or make no prediction. Method 1 slightly lags in ``both labels correct'', but excels in minimizing ``both labels incorrect'' cases. Method 2 performs best in ``both labels correct'' but has more ``both labels incorrect'' instances, indicating more entirely incorrect predictions. Method 3 has a higher ``both labels incorrect'' count, but it can consistently predict labels for all instances.

\begin{table*}[h]
\footnotesize
\centering
\begin{tabular}{lcccc}
\toprule
Method & Both Labels Correct & One Label Correct & Both Labels Incorrect & No Prediction \\
\midrule
Method 1 & 363 (39\%) & 404 (43\%) & 134 (14\%) & 41 (4\%) \\
Method 2 & 392 (42\%) & 343 (36\%) & 168 (18\%) & 39 (4\%) \\
Method 3 & 382 (40\%) & 346 (37\%) & 214 (23\%) & 0 (0\%) \\

\bottomrule
\end{tabular}
\caption{A comparison of methods on the predictions for multi-label examples (examples annotated with two labels).}
\label{tab:method_comparison}
\end{table*}

\begin{table*}[]
\footnotesize
    \centering
\begin{tabular}{lllll}
\toprule
          Label & Single(base)& Multi(base) & Single(large) & Multi (large) \\
\midrule
     Concession &        47.08$\pm$2.69 &      51.99$\pm$4.3 & 61.17$\pm$4.07 &      61.2$\pm$3.93 \\
       Contrast &        49.01$\pm$2.32 &     52.94$\pm$2.25 & 57.19$\pm$4.07 &     59.76$\pm$2.55 \\
          Cause &        66.31$\pm$1.75 &      66.0$\pm$1.56 &70.89$\pm$1.52 &     70.44$\pm$1.61 \\
   Cause+Belief &         4.13$\pm$5.74 &      6.52$\pm$9.08 & 8.59$\pm$6.99 &    10.06$\pm$11.74 \\
      Condition &        78.88$\pm$8.87 &     80.16$\pm$7.84 &       84.91$\pm$10.74 &    84.55$\pm$10.21 \\
        Purpose &        91.34$\pm$2.56 &     91.47$\pm$1.91 &        92.04$\pm$2.75 &     92.53$\pm$2.68 \\
    Conjunction &        61.21$\pm$2.56 &     63.52$\pm$2.93 &        68.03$\pm$1.73 &     67.48$\pm$2.98 \\
    Equivalence &        15.75$\pm$8.81 &     16.67$\pm$8.38  &        22.56$\pm$12.0 &     25.85$\pm$7.04 \\
  Instantiation &        56.63$\pm$7.57 &     60.86$\pm$4.91 &        63.62$\pm$3.68 &      61.3$\pm$4.62 \\
Level-of-detail &        53.61$\pm$2.97 &     54.07$\pm$3.33  &        58.58$\pm$2.83 &      58.1$\pm$1.62 \\
         Manner &        79.86$\pm$9.26 &    77.15$\pm$10.21 &       80.12$\pm$10.92 &    77.35$\pm$12.36 \\
   Substitution &       60.34$\pm$13.13 &     65.67$\pm$8.61&        70.34$\pm$6.17 &     71.92$\pm$8.45 \\
   Asynchronous &         61.6$\pm$4.25 &     60.93$\pm$3.97 &        68.12$\pm$2.97 &     67.93$\pm$3.96 \\
    Synchronous &        41.5$\pm$12.83 &     35.46$\pm$7.96 &        40.93$\pm$8.95 &     46.51$\pm$11.4 \\
    \hline
          Total &         54.8$\pm$1.85 &     55.96$\pm$0.84 &        60.51$\pm$1.32 &     61.07$\pm$1.64 \\
\bottomrule
\end{tabular}
\caption{Comparative evaluation of cross-validation Macro-F1 scores for multi-label versus single-label prediction methods, with multi-label predictions assessed using single-label evaluation criteria. This study employs RoBERTa$_{\text{base}}$  and RoBERTa$_{\text{large}}$  for comparisons, providing standard deviations over 12-fold. ``Total'' here refers to the average scores for all labels.}
    \label{tab:my_label}
\end{table*}

\subsection{Multi-Label vs. Single-Label Prediction: A Comparative Performance Analysis}
To compare multi-label and single-label prediction methods, we evaluated Method 2 under single-label criteria. In this evaluation, the highest probability label is chosen, and for multi-label examples, we consider it correct if the predicted label matches one of the gold labels. We did not evaluate the single-label prediction method in terms of multi-label criteria since it is not feasible when the method inherently provides only a single prediction. If we use the top two largest probabilities for the single prediction methods, it would be assumed that all instances has two labels.


We utilize RoBERTa to obtain the [CLS] representation for each example for both single label prediction method and Method 2, but Method 2 uses separate classification heads for binary classification per class, while single-label classification employs a multi-class mapping layer,  with a size of $\mathbb{R}^{h \times |C|}$.
Specifically, num\_class takes the value of 14 here as the number of the labels is 14. Training loss for both methods is softmax cross-entropy. We use both RoBERTa$_{\text{base}}$ and RoBERTa$_{\text{large}}$ for comparisons.

The results in Table 4 indicate that, while the evaluation method for both single classification methods and multi-label classification methods is the same, based on the single-label evaluation criteria, the multi-label prediction method outperforms the single-label prediction method for both RoBERTa$_{\text{base}}$ and RoBERTa$_{\text{large}}$. This suggests that multi-label prediction does not compromise the performance of single-label prediction.

It should also be noted that multi-label classification methods do not necessarily increase computational complexity, since they reduce model redundancy. That is, instead of creating a separate training example for each sense of a multi-label token, a multi-label approach relies on a single model that can
predict multiple labels simultaneously. Moreover, multi-label methods are not necessarily more complicated,
using (in our experiments) no more computational resources than the single-label prediction method.

\section{Fine-grained Analysis of the Experimental results}

\begin{figure}
\centering
\subfigure[]{
\begin{minipage}[t]{\linewidth}
\centering
\includegraphics[width=6cm,height=4cm]{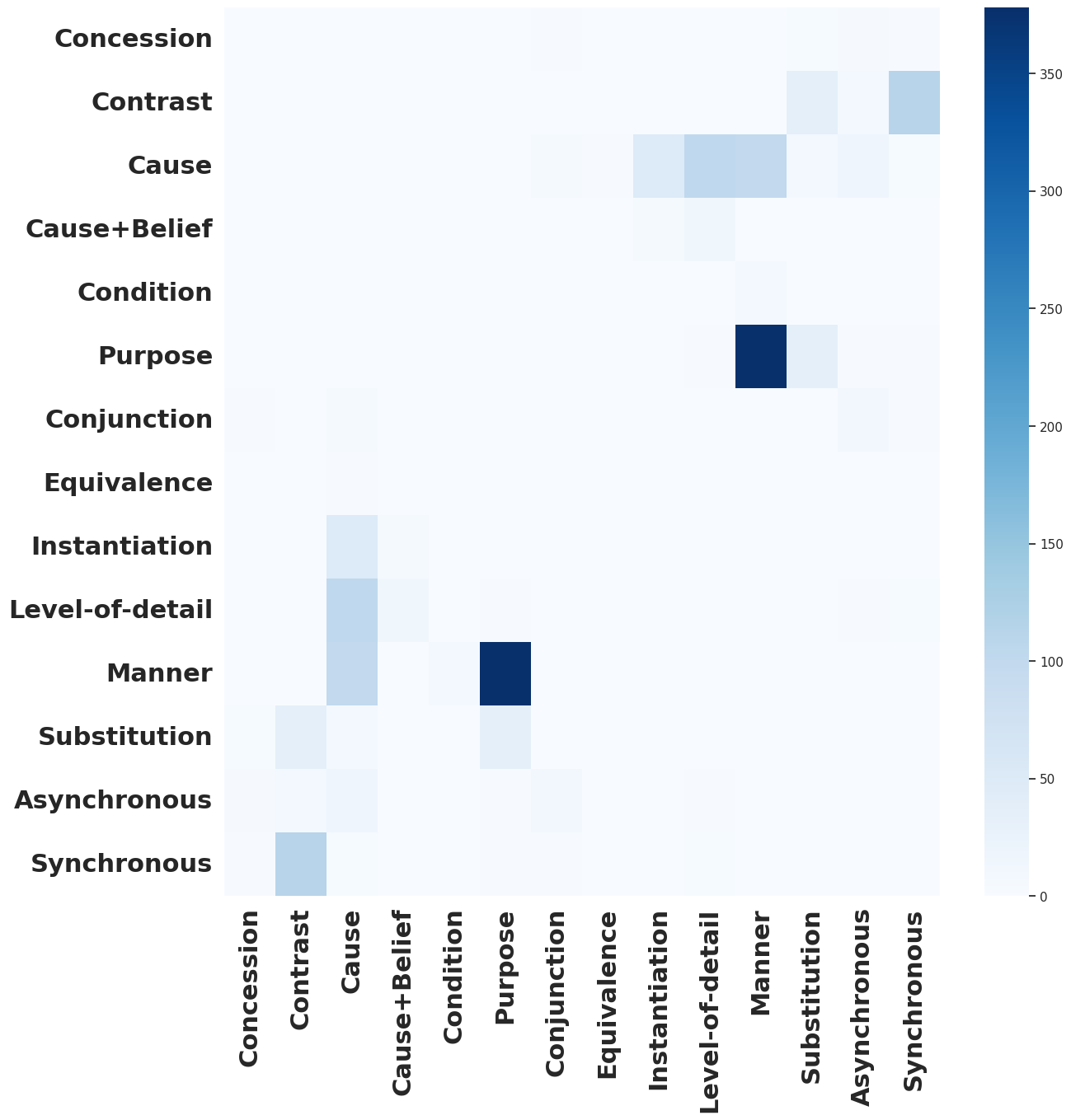}
\label{fig1}
\end{minipage}%
}

\subfigure[]{
\begin{minipage}[t]{\linewidth}
\centering
\includegraphics[width=6cm,height=4cm]{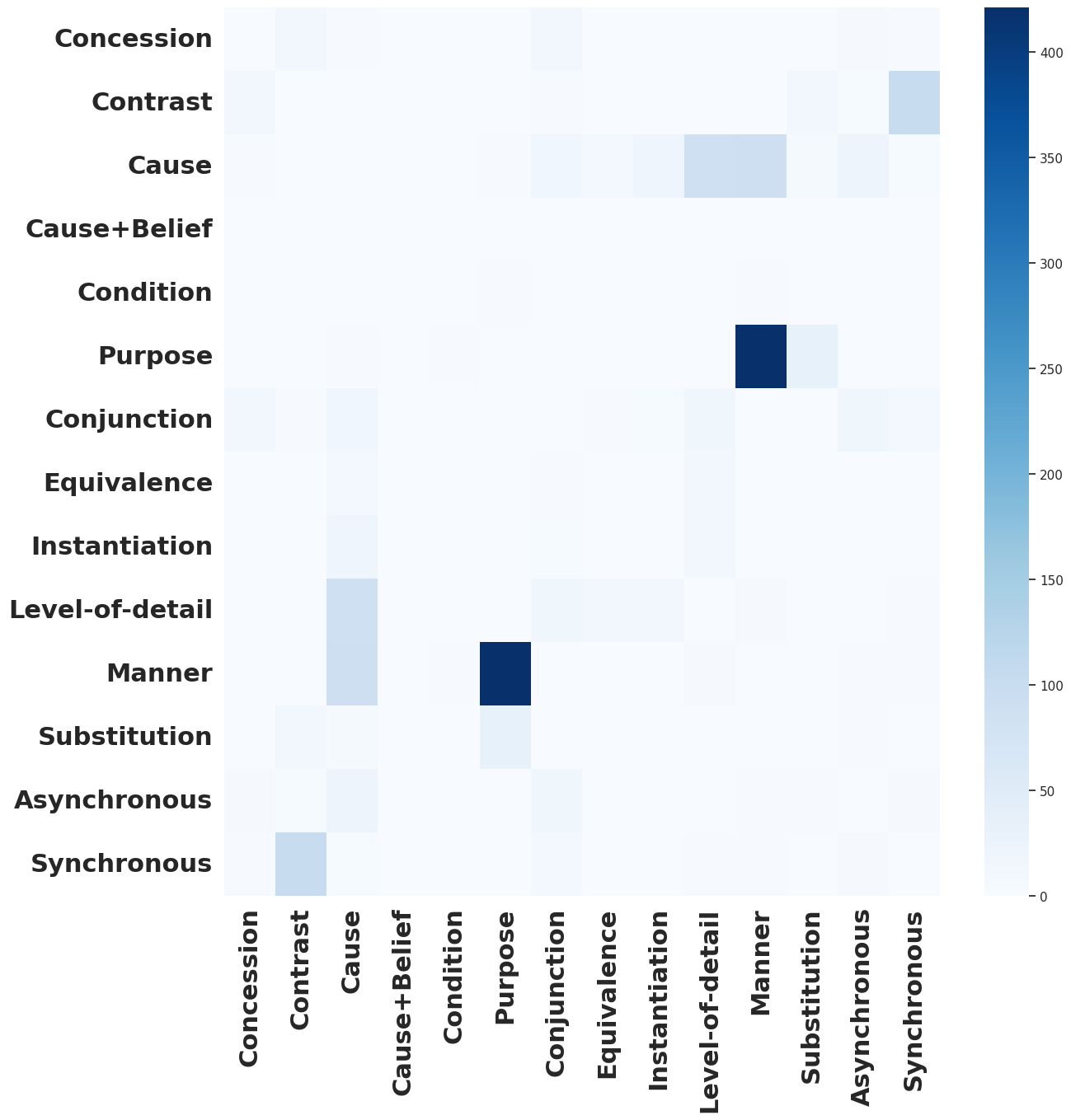}
\label{fig2}
\end{minipage}%
}
\caption{Co-occurrence of label pairs in the dataset and in the prediction. The upper sub-figure is for the gold label pairs, while the lower is for the predicted pairs.} 
\end{figure}



\begin{figure*}
\centering
\includegraphics[width=14cm, height = 10cm]{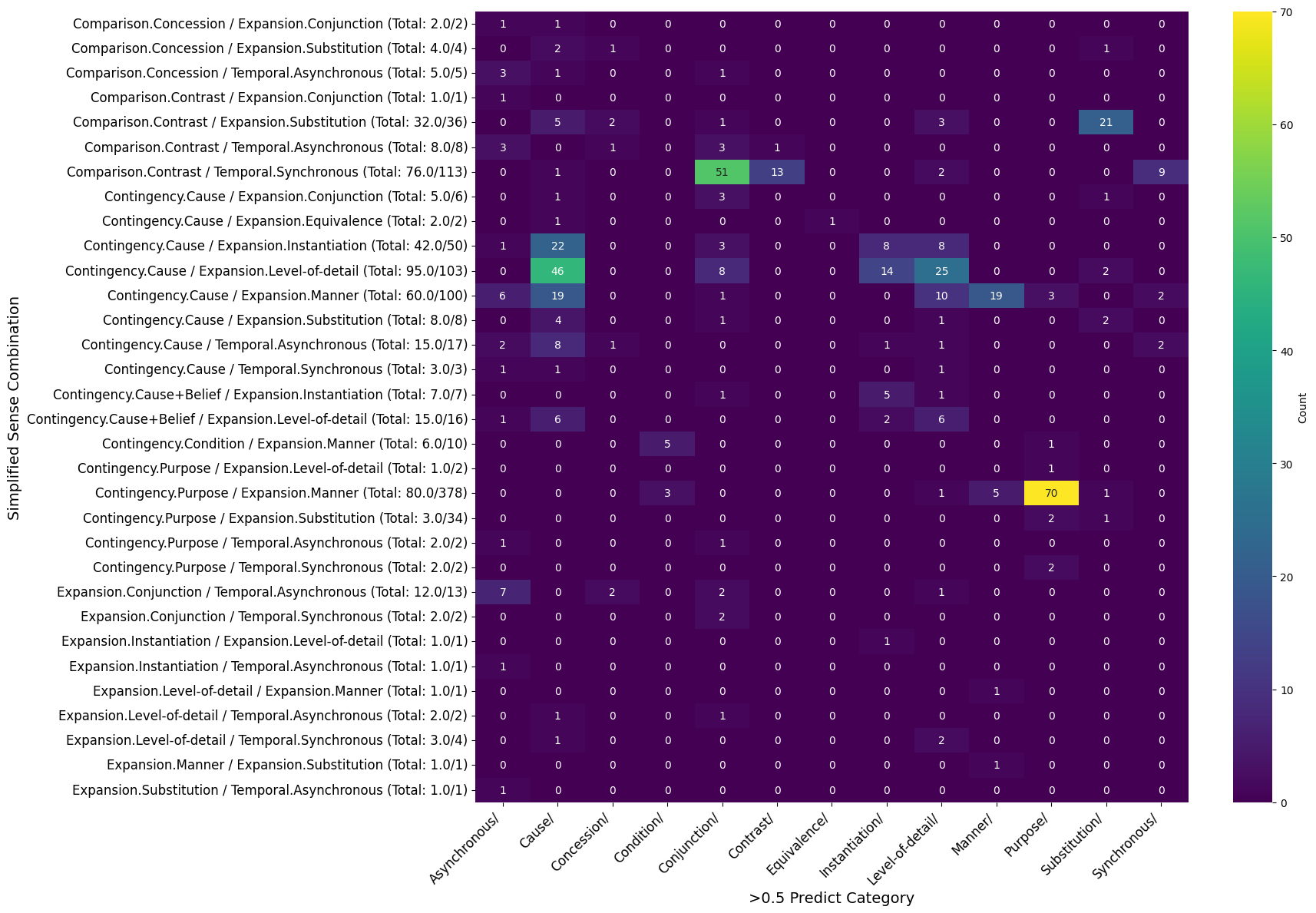}
\caption{Heatmap of underpredicted multi-label instances. This figure displays the distribution of instances where two labels are annotated but only one is predicted.}

\end{figure*}

\subsection{The Multi-label Classification Method Can Capture the Label Correlations}
In Figure 1, two co-occurrence matrices visualize the joint distribution of label pairs in the dataset and in the prediction. The upper one is the co-occurrence of label pairs in the dataset and the lower one is the co-occurrence of the predicted label pairs. The gold label pairs are the multi-label data in the dataset. 

For the predicted label pairs, all the predictions which have two labels on the test sets over 12-fold are used. Darker shades represent a higher frequency or probability of the label pairs co-occurring. In Method 2, for each label, the model independently predicts whether the label is the gold label by doing a binary classification. Despite that, the two matrices look very similar, which indicates that the model seems to have implicitly captured the correlations between the labels from the data. 

However, in Table 3, we have knew that method 2 predict two labels correctly for only 42\% of multi-label examples. This implies that the model's capacity to capture label correlations does not inherently guarantee very high accuracy in predicting multi-labels. For example, although the model can capture the correlations between ``Purpose'' and ``Manner'', it cannot distinguish which cases have both labels from those cases which only have ``Purpose'' or ``Manner''.

\subsection{When Multi-label Examples Are Predicted as Single Label}
Figure 2 presents a heatmap, which depicts instances where only one label is predicted for multi-label examples(with two annotated labels). Vertical axes correspond to gold labels, while horizontal ones correspond to predicted labels. Numbers beside the vertical axis show how many examples had two labels but received only one predicted label and the number of such label pairs in the dataset.

\textbf{Challenge in ``Purpose\&Manner'' vs. ``Purpose''}
Figure 2 reveals around a quarter of instances that should have both ``Purpose'' and ``Manner'' labels are only labeled as``Purpose''. Moreover, approximately one-third of instances labeled as ``Purpose'' are predicted as both ``Purpose'' and ``Manner''. 

The following are two examples for ``Purpose\&Manner'' vs. ``Purpose'' :

 \begin{enumerate}
    \item [(1)]
[they exercise]$_{1}$ [to lose weight.]$_{2}$. Labels: Purpose and Manner
\end{enumerate} 

In Example (1), the purpose of exercising is to ``lose weight'', while the manner in which weight loss is to be achieved is through exercising. Thus, both ``Purpose'' and ``Manner''  are appropriate sense labels.

\begin{enumerate}
    \item [(2)]
[Mr. Achenbaum will work with clients]$_{1}$ [to determine the mix of promotion, merchandising, publicity and other marketing outlets, and to integrate those services]$_{2}$. Label: Purpose 

\end{enumerate}

In contrast, in Example (2), while the purpose of working with clients is to determine their service needs, the annotators appear to have decided that working with clients is not the manner by which their service needs are determined. As such, only ``Purpose'' was annotated as a sense label.

While one might disagree with the annotators' labelling decisions here, the value of multi-sense prediction is to highlight potentially questionable cases that might well justify further review \cite{klie-etal-2023-annotation}.

The two examples indicate the challenges the model faces in distinguishing between ``Purpose and Manner'' and ``Purpose'' in certain cases, despite similar linguistic structures. More work is needed to determine when an example demonstrates both `Purpose and Manner''versus just ``Purpose''.



\textbf{Challenge in rare label combinations}
The model encounters difficulties with infrequent label combinations, often predicting just one label or even assigning an incorrect label unrelated to the combination. PDTB exhibits imbalances not only in single-label data but also in multi-label data. Some label combinations are more prevalent than others, potentially hindering the model's ability to generalize effectively to those less common combinations.


\textbf{Challenge in identifying both elaborative and argumentative discourse relations}
\citet{Scholman2017ExamplesAS} find that annotators often disagree on whether the relations are elaborative(illustrate / specify a situation)or argumentative(serve as an argument for a claim), and they note that the relation ``Example'' and the relation ``Specifications'' in PDTB-2 can embody both. It appears to be the case that annotators do not appear to have annotated similar relations consistently in the PDTB-3. That is, Figure 1 shows that there are some instances of co-occurrence, but it is only a small
fraction of cases where multiple labels might be justified. In the PDTB annotation, the annotators are not required to annotate both elaborative and argumentative relations simultaneously, which could be causing confusion for models in classification tasks. Therefore, it would probably be worth adding this multi-label annotation to the corpus.

\section{Other Investigations}

\subsection{Replacing Cross Entropy Loss with Focal Loss}
Focal Loss, an adaptation of Cross-Entropy Loss (CE), addresses class imbalance by emphasizing challenging examples. Initially designed for object detection in computer vision by \cite{Lin2017FocalLF}, it has been applied in recent NLP studies \cite{tan-etal-2022-document,wang-etal-2022-calibrating}. Our dataset, PDTB-3, exhibits imbalances in single-label and multi-label data. While single-label tasks often use standard cross-entropy loss for IDRR, we explored focal loss for the multi-label classification in IDRR.

Focal Loss reduces weights for well-classified instances and accentuates challenging ones, modulating the loss for confidently predicted instances. For positive samples (y = 1), it is defined as::

The focal loss function is defined as follows:
for positive samples $(y=1)$:
\begin{equation} L_{fl}(p) = (1 - p)^\gamma \log(p) \end{equation}
For negative samples $(y=0)$:
\begin{equation} L_{fl}(p) = p^\gamma \log(1 - p) \end{equation}

Here, $\gamma$ serves as the focusing parameter, controlling the rate at which easy instances are down-weighted. We set it to 1 for positive samples, and 4 for negative.

In our implementation, we replaced the cross-entropy loss in Method 2 with the aforementioned focal loss function without further modifications. Table 5 presents the outcomes of employing both loss functions, demonstrating that the adoption of focal loss enhances the overall performance when applied to Method 2 in the context of IDRR.

\begin{table}
\footnotesize
\centering
\begin{tabular}{lllll}
\toprule
          Label & Cross-entropy & Focal loss\\
\midrule
     Concession &        50.64$\pm$3.97 &           52.13$\pm$4.56     \\
       Contrast &           49.29$\pm$5.37 &         49.8$\pm$3.7     \\
          Cause &        65.13$\pm$2.13 &            67.11$\pm$2.57  \\
   Cause+Belief &           0.0$\pm$0.0 &               9.88$\pm$4.32 \\
      Condition &        77.84$\pm$7.37 &         78.39$\pm$9.18  \\
        Purpose &         92.41$\pm$2.4 &          92.34$\pm$2.26 \\
    Conjunction &        63.21$\pm$3.16 &         64.15$\pm$3.13  \\
    Equivalence &       17.55$\pm$10.12 &             23.69$\pm$7.16 \\
  Instantiation &        59.76$\pm$5.27 &          58.93$\pm$4.89  \\
Level-of-detail &        52.84$\pm$3.38 &            54.05$\pm$1.79  \\
         Manner &       58.94$\pm$10.65 &          58.53$\pm$12.22  \\
   Substitution &         62.83$\pm$9.2 &           62.26$\pm$7.69  \\
   Asynchronous &        62.19$\pm$3.24 &           60.73$\pm$4.73  \\
    Synchronous &         30.3$\pm$7.85 &            30.33$\pm$5.35 \\
    \hline
          Total &        53.16$\pm$1.39 &           54.45$\pm$1.19 \\
\bottomrule
\end{tabular}
\caption{Comparative analysis of F1 scores for Method 2 using RoBERTa$_{\text{base}}$: evaluating cross-entropy and focal loss functions over 12-fold cross-validation with reported standard deviations. ``Total'' here refers to the average scores for all labels.}

\end{table}

\subsection{Replacing Section-level Cross Validation with Example-Level}
Splitting the data at section-level may not be optimal for multi-label classification, since the multi-label examples are limited in number and are not evenly distributed across sections. Therefore, alternative strategies can be considered.

We tried an example-level method to offer a better mix of examples to train more robust models, especially when dealing with sparse labels. We first separated the multi-label data from the single-label data. Then, we divided the multi-label data into 12 portions, and the single-label data was also divided into 12 portions, with each portion having the same proportion of the number of each label. Next, we combined one of the 12 portions of multi-label data with one of the 12 portions of single-label data to obtain a merged set of 12 data portions. Finally, we randomly selected one portion as the test set, another portion as the validation set, and the remaining 10 portions as the training set, thus creating 12 folds of cross-validation data.

\begin{table*}[htp]
\small
\centering
\begin{tabular}{lccc}
\toprule
          Label & Method 1 & Method 2 & Method 3 \\
\midrule
     Concession &        45.73$\pm$3.06 &        48.54$\pm$4.14 &        49.09$\pm$4.27 \\
       Contrast &        48.42$\pm$3.42 &        50.05$\pm$3.92 &        50.02$\pm$3.71 \\
          Cause &        63.78$\pm$1.96 &        64.59$\pm$1.86 &        64.69$\pm$1.93 \\
   Cause+Belief &           0.00$\pm$0.00 &           0.00$\pm$0.00 &         1.02$\pm$3.57 \\
      Condition &        72.26$\pm$9.37 &        75.61$\pm$8.65 &        77.01$\pm$8.95 \\
        Purpose &        92.51$\pm$1.99 &        92.79$\pm$1.84 &         92.7$\pm$1.98 \\
    Conjunction &        61.57$\pm$2.48 &        62.26$\pm$2.76 &         62.6$\pm$2.42 \\
    Equivalence &        12.94$\pm$5.24 &         17.2$\pm$8.71 &        16.08$\pm$8.08 \\
  Instantiation &        58.96$\pm$4.55 &         59.64$\pm$4.4 &         59.82$\pm$4.2 \\
Level-of-detail &        51.34$\pm$3.02 &        51.99$\pm$2.87 &        51.93$\pm$3.01 \\
         Manner &        57.61$\pm$5.45 &        59.01$\pm$5.78 &       46.91$\pm$18.12 \\
   Substitution &        60.38$\pm$6.47 &        62.09$\pm$6.32 &        59.95$\pm$7.18 \\
   Asynchronous &        60.59$\pm$3.91 &        61.75$\pm$3.65 &        61.41$\pm$3.74 \\
    Synchronous &        29.04$\pm$5.91 &        28.67$\pm$6.86 &        29.29$\pm$6.96 \\
    \hline
          Total &        51.08$\pm$0.88 &        52.44$\pm$1.72 &        51.61$\pm$1.96 \\
\bottomrule
\end{tabular}
\caption{Example level: comparison of Macro-F1 scores with standard deviations detailed across three methods using RoBERTa$_{\text{base}}$. Note: the methods in Table 1 and this table are consistent, differing only in the application of section-level versus example-level cross-validation techniques. ``Total'' here refers to the average scores for all labels.}
\end{table*}

Table 6 shows the results for three methods where the cross-validation is done at example-level. We can compare Table 6 with Table 1. Method 2 stands out with the highest total F1 score in both cross-validation approaches. Method 2 also shows less variability for example-level. However, we leave it to future work to decide which cross-validation strategy should be followed.

\section{Discussions: Key Insights and Future Study}
\textbf{Embracing Multi-Label Classification in Discourse Relation Recognition}
This study not only identifies the limitations of single-label prediction but also explores various multi-label classification methods. The results demonstrate that multi-label classification methods can effectively predict multiple labels and capture label correlations. Interestingly, the straightforward multi-label classification approach outperforms single-label prediction under single-label criteria. We advocate for the broader use and further development of multi-label prediction methods in this domain, supported by our motivational insights and the practical results we have obtained.


\noindent\textbf{Navigating the Challenges of Multi-Label Scenarios}
Compared to traditional single-label predictions, treating discourse relation recognition as a multi-label prediction task introduces more challenges to this task. 

Firstly, the model must accurately determine whether an example should be treated as a single-label or multi-label instance; for instance, distinguishing between cases like ``Purpose and Manner'' and ``Purpose'' alone. Furthermore, the model needs to more effectively capture and exploit the inter-label relationships to enhance multi-label classification performance. For example, it could be better that the model can identify both elaborative and argumentative relations for certain examples, rather than disregarding one in favor of the other. Thirdly, the issue of data imbalance persists not only in single-label data but also in multi-label data, making it more challenging to address the problem of imbalanced data.

\noindent\textbf{Diversifying and Expanding Multi-Label Discourse Datasets}
Based on our experimental analysis, we have observed that the model often struggles to accurately predict certain rare combinations of labels. The availability of annotated multi-label examples in PDTB-3 is limited, with some label pairs being exceptionally infrequent in this dataset. 

Moreover, PDTB primarily consists of news discourse, while it has been noted that the text genre significantly impacts the distribution of discourse relations\cite{scholman-etal-2022-discogem}. By confining the dataset to news texts, there is a missed opportunity to comprehensively understand and model discourse relations and label associations across a wider range of text types.

Additionally, \citet{rohde-etal-2018-discourse} pointed out that while other works on discourse coherence acknowledge the possibility of multiple relations between discourse segments, they typically lack annotation for more than one discourse relation. For example, Rhetorical Structure Theory (RST) allows for multiple possible analyses but tends to select one that aligns with the writer's goals \cite{Mann1989RhetoricalST}. Therefore, the need for more diverse and abundant multi-label data is crucial to enable systems to learn more effectively and comprehensively.

\noindent\textbf{Exploring Advanced Modeling Techniques}
In this study, we applied three foundational multi-label classification methods to IDRR, and we have found that the focal loss is a superior choice for the multi-label classification method in this task. Nevertheless, addressing the complexities of discourse relations, data imbalance, and the limited availability of annotated multi-discourse relations,etc. requires more advanced customized modeling techniques.

\noindent\textbf{Extend the Multi-label Classification Methods to Other Types of Discourse Relations}
In addition to annotating multiple sense relations holding in the PDTB when no discourse connective was present, annotators could also record a distinct implicit relation holding in the context of an explicit connective that doe not itself convey that relation (\cite{webber2019penn} Section 3.4). This approach allowed them to represent relations inferred not just from explicit connectives but also from the underlying arguments themselves. While previous work on discourse relation recognition has ignored such cases, it would be valuable to test our multi-label classification methods on these cases as well.

In summary, the application of multi-label methods to implicit discourse relation recognition is valuable and feasible, yet there is great room for further improvement both in terms of data and methodology.

\section{Conclusions} 
We conducted a comprehensive exploration of implicit discourse relation recognition (IDRR) as a multi-label classification problem, addressing real-world complexity and label interdependence. We found that multi-label approaches can have better performance than traditional single-label metrics when evaluated on the criteria for the single-label prediction. 

Our exploration of multi-label frameworks, loss function optimization, and cross-validation strategies sets the stage for future research advancements. Additionally, our detailed analysis can provide valuable insights for further investigations in this field.

\section*{Acknowledgments}
This work was supported in part by the UKRI Centre for Doctoral Training in Natural Language Processing, funded by the UKRI (grant EP/S022481/1), the University of Edinburgh. The authors also gratefully acknowledge the University of Edinburgh and Huawei Joint Laboratory for their support. 

\section*{Limitations}
In our work, we only used PDTB-3 as our evaluation dataset, despite the scarcity of datasets annotating multiple discourse relations. However, some other datasets such as GUM \cite{Zeldes2017} and DiscoGem \cite{scholman-etal-2022-design} probably can be considered for our evaluation and analysis. DiscoGem \cite{scholman-etal-2022-design} can be considered to see whether it either agree with or contradict multi-sense PDTB-3 sense annotation, although DiscoGeM is not inherently a multi-label dataset but rather a collection with diverse annotations stemming from varying perspectives.

\bibliography{anthology,custom}
\bibliographystyle{acl_natbib}
\newpage
\appendix

\section{Dataset Statistics}
Table 7 and Table 8 give the statistics of the single-label data and multi-label data. Both distributions are very imbalanced. For Table 8, the counts for Level-2 label pairs vary widely, ranging from as low as 4 instances to as high as 378 instances. Some Level-2 label pairs, such as ``Purpose/Manner'' and ``Cause/Level-of-detail'', have relatively high counts, suggesting a significant presence of these combinations in the dataset. Others, like ``Asynchronous/Contrast'' and ``Concession/Asynchronous'', have lower counts, indicating less frequent occurrences of these specific pairs. In summary, the distribution of discourse relations and the label pairs in the PDTB 3.0 dataset varies, with some categories and pairs being more prevalent than others. 

\begin{table}[t]
\footnotesize
\centering
\begin{tabular}{@{}lr@{}}
\toprule
Label & \( n \) \\
\midrule
Comparison & 2518 \\
Contingency & 7583 \\
Expansion & 10833 \\
Temporal & 1828 \\
\hline
Comparison.Concession & 1494 \\
Comparison.Contrast & 983 \\
Contingency.Cause & 5785 \\
Contingency.Cause+Belief & 202 \\
Contingency.Condition & 199 \\
Contingency.Purpose & 1373 \\
Expansion.Conjunction & 4386 \\
Expansion.Equivalence & 336 \\
Expansion.Instantiation & 1533 \\
Expansion.Level-of-detail & 3361 \\
Expansion.Manner & 739 \\
Expansion.Substitution & 450 \\
Temporal.Asynchronous & 1289 \\
Temporal.Synchronous & 539 \\
\bottomrule
\end{tabular}
\caption{Label counts for level-1 senses and level-2 senses that have more than 100 annotated instances in PDTB-3.}

\label{tab:my_label}
\end{table}

\begin{table}

\begin{tabular}{lr}
\toprule
                         Label &  number \\
\midrule
           Asynchronous/Cause &       6 \\
        Asynchronous/Contrast &       4 \\
   Cause+Belief/Instantiation &       7 \\
 Cause+Belief/Level-of-detail &      13 \\
           Cause/Asynchronous &      11 \\
          Cause/Instantiation &      50 \\
        Cause/Level-of-detail &     101 \\
                 Cause/Manner &     100 \\
           Cause/Substitution &       7 \\
      Concession/Asynchronous &       4 \\
      Concession/Substitution &       4 \\
             Condition/Manner &      10 \\
     Conjunction/Asynchronous &      12 \\
            Conjunction/Cause &       4 \\
        Contrast/Asynchronous &       4 \\
        Contrast/Substitution &      35 \\
               Purpose/Manner &     378 \\
         Purpose/Substitution &      34 \\
         Synchronous/Contrast &     112 \\
  Synchronous/Level-of-detail &       4 \\
\bottomrule
\end{tabular}
\caption{Label counts for level-2 sense pairs that have more than 3 annotated instances in PDTB-3.}
\vspace{-0.4cm}

\end{table}

\section{More Results for Three Methods}
\begin{table*}[htbp]
\small
    \centering
    \begin{tabular}{cccccccccc}
        \toprule
        \textbf{Label} & \textbf{Precision $\pm$ Std} & \textbf{Recall $\pm$ Std} & \textbf{F1 $\pm$ Std} & \textbf{Hamming Loss $\pm$ Std} \\
        \midrule
        Concession  &49.75$\pm$6.85&53.07$\pm$6.82& 50.98 $\pm$ 5.06 & - \\
        Contrast &46.37 $\pm$ 4.50& 56.72$\pm$7.63 & 50.58 $\pm$ 3.40 & - \\
        Cause &66.75 $\pm$ 2.80 &64.64$\pm$3.57& 65.57 $\pm$ 1.76  & - \\
        Cause+Belief&0.00 $\pm$ 0.00 &0.00$\pm$0.00  & 0.00 $\pm$ 0.00  & - \\
        Condition& 72.60 $\pm$ 5.29&80.36$\pm$9.84  & 75.99 $\pm$ 5.95 & - \\
        Purpose & 92.49 $\pm$ 2.39& 92.55$\pm$2.44& 92.50 $\pm$ 2.01  & - \\
        Conjunction & 59.29 $\pm$ 6.16 &65.95$\pm$4.13 & 62.12 $\pm$ 3.05 & - \\
        Equivalence & 10.07 $\pm$ 6.16 &21.93$\pm$13.32 & 12.99 $\pm$ 7.56  & - \\
        Instantiation & 58.70 $\pm$ 4.58 &59.98$\pm$8.22 & 58.76 $\pm$ 4.02 & - \\
        Level-of-detail& 49.35 $\pm$ 6.15 & 53.14$\pm$3.83  & 50.93 $\pm$ 3.97 & -  \\
        Manner& 68.81 $\pm$ 10.06 &52.59$\pm$13.40 & 58.76 $\pm$ 11.39  & -  \\
        Substitution & 64.12 $\pm$ 12.21 &65.55$\pm$12.30 & 64.11 $\pm$ 10.35 & -  \\
        Asynchronous & 61.84 $\pm$ 5.59 &63.73$\pm$6.54  & 62.46 $\pm$ 4.01&-  \\
        Synchronous  & 24.08 $\pm$ 9.49 &34.47$\pm$12.87& 27.40 $\pm$ 9.48& -  \\
        \hline
        Total & 51.73 $\pm$ 1.91 &54.62$\pm$1.92 & 52.37 $\pm$ 1.62  & 5.73$\pm$0.22\\
        \bottomrule
    \end{tabular}
    \caption{Results for Method 1: Level-2 implicit discourse relation recognition cross-validation results on PDTB-3.``Total'' here refers to the average scores for all labels.}
\end{table*}

\begin{table*}[ht]
\small
\centering
\label{tab:results}
\begin{tabular}{lcccc}
\toprule
Label & Precision$\pm$ Std & Recall $\pm$Std & F1 $\pm$ Std & Hamming Loss $\pm$ Std \\
\midrule
Concession & 53.78 $\pm$ 8.46 & 50.35 $\pm$ 5.04 & 51.59 $\pm$ 4.61 & - \\
Contrast & 57.20 $\pm$ 6.65 & 46.55 $\pm$ 5.12 & 50.82 $\pm$ 2.99 & - \\
Cause & 66.79 $\pm$ 4.25 & 63.96 $\pm$ 3.71 & 65.15 $\pm$ 1.74 & - \\
Cause+Belief & 0.00 $\pm$ 0.00 & 0.00 $\pm$ 0.00 & 0.00 $\pm$ 0.00 & - \\
Condition & 82.41 $\pm$ 9.78 & 80.29 $\pm$ 8.83 & 80.97 $\pm$ 7.51 & - \\
Purpose & 92.57 $\pm$ 2.56 & 92.83 $\pm$ 2.77 & 92.68 $\pm$ 2.34 & - \\
Conjunction & 64.58 $\pm$ 3.99 & 62.95 $\pm$ 7.78 & 63.32 $\pm$ 3.34 & - \\
Equivalence & 29.68 $\pm$ 25.37 & 10.93 $\pm$ 6.43 & 14.55 $\pm$ 8.43 & - \\
Instantiation & 58.84 $\pm$ 8.08 & 61.28 $\pm$ 5.62 & 59.67 $\pm$ 5.73 & - \\
Level-of-detail & 53.81 $\pm$ 3.80 & 50.85 $\pm$ 7.69 & 51.80 $\pm$ 3.80 & - \\
Manner & 53.36 $\pm$ 15.81 & 67.77 $\pm$ 9.88 & 58.60 $\pm$ 13.47 & - \\
Substitution & 63.79 $\pm$ 10.45 & 61.70 $\pm$ 8.20 & 62.35 $\pm$ 7.83 & - \\
Asynchronous & 64.09 $\pm$ 6.87 & 60.96 $\pm$ 5.58 & 62.10 $\pm$ 3.88 & - \\
Synchronous & 34.30 $\pm$ 13.41 & 29.30 $\pm$ 9.22 & 30.28 $\pm$ 7.30 & - \\
\hline
Total & 55.37 $\pm$ 2.35 & 52.84 $\pm$ 0.90 & 53.13 $\pm$ 0.92 & 5.70 $\pm$ 0.25 \\
\bottomrule
\end{tabular}
\caption{Results for Method 2: Level-2 implicit discourse relation recognition cross-validation results on PDTB-3.``Total'' here refers to the average scores for all labels.}
\end{table*}

\begin{table*}[ht]
\small
\centering
\label{tab:results}
\begin{tabular}{lcccc}
\toprule
Label & Precision $\pm$ Std & Recall $\pm$ Std& F1 $\pm$ Std & Hamming Loss $\pm$ Std \\
\midrule
Concession & 50.73 $\pm$ 6.40 & 51.97 $\pm$ 5.28 & 50.86 $\pm$ 2.91& -  \\
Contrast & 47.14 $\pm$ 6.44 & 50.94 $\pm$ 5.20 & 48.49 $\pm$ 3.46 & - \\
Cause & 63.69 $\pm$ 3.67 & 67.34 $\pm$ 3.59 & 65.34 $\pm$ 2.19 & - \\
Cause+Belief & 12.78 $\pm$ 28.21 & 1.93 $\pm$ 3.66 & 3.04 $\pm$ 5.58& -  \\
Condition & 79.40 $\pm$ 11.78 & 77.24 $\pm$ 10.08 & 78.01 $\pm$ 10.00& -  \\
Purpose & 92.32 $\pm$ 3.23 & 92.90 $\pm$ 2.21 & 92.58 $\pm$ 2.21 & - \\
Conjunction & 63.45 $\pm$ 5.96 & 61.75 $\pm$ 6.81 & 62.11 $\pm$ 3.08 & - \\
Equivalence & 23.47 $\pm$ 10.20 & 15.47 $\pm$ 6.18 & 17.85 $\pm$ 6.42 & - \\
Instantiation & 56.78 $\pm$ 9.55 & 61.83 $\pm$ 5.72 & 58.49 $\pm$ 6.15 & - \\
Level-of-detail & 50.70 $\pm$ 3.06 & 51.92 $\pm$ 4.74 & 51.09 $\pm$ 2.26 & - \\
Manner & 15.06 $\pm$ 6.09 & 55.07 $\pm$ 12.14 & 23.23 $\pm$ 7.94& -  \\
Substitution & 50.23 $\pm$ 11.18 & 60.61 $\pm$ 8.68 & 54.46 $\pm$ 9.18& -  \\
Asynchronous & 62.45 $\pm$ 5.65 & 60.46 $\pm$ 5.06 & 61.20 $\pm$ 3.72 & - \\
Synchronous & 31.46 $\pm$ 7.93 & 30.31 $\pm$ 8.43 & 30.30 $\pm$ 6.96 & - \\
\hline
Total & 49.98 $\pm$ 1.50 & 52.84 $\pm$ 0.99 & 49.79 $\pm$ 1.12 & 5.98 $\pm$ 0.22 \\
\bottomrule
\end{tabular}
\caption{Results for Method 3: Level-2 implicit discourse relation recognition cross-validation results on PDTB-3.}

\end{table*}

\begin{table}
\small
\begin{tabular}{lll}
\toprule
          Label & Cross-entropy(large)& Focal loss(large) \\
\midrule
     Concession &         59.5$\pm$2.49 &     61.59$\pm$4.87 \\
       Contrast &        57.32$\pm$3.98 &     56.92$\pm$2.94 \\
          Cause &        70.28$\pm$1.45 &     71.75$\pm$1.45 \\
   Cause+Belief &          8.47$\pm$5.3 &      8.92$\pm$8.28 \\
      Condition &        84.41$\pm$6.84 &     84.89$\pm$9.86 \\
        Purpose &        92.25$\pm$2.26 &      92.97$\pm$2.5 \\
    Conjunction &        67.79$\pm$2.16 &     66.68$\pm$3.51 \\
    Equivalence &       28.91$\pm$10.07 &      32.29$\pm$7.7 \\
  Instantiation &        63.23$\pm$2.33 &     60.46$\pm$5.67 \\
Level-of-detail &        57.56$\pm$1.81 &     57.93$\pm$1.71 \\
         Manner &        53.6$\pm$12.61 &    55.11$\pm$14.34 \\
   Substitution &        71.62$\pm$11.3 &     70.54$\pm$9.83 \\
   Asynchronous &         66.1$\pm$3.96 &     64.59$\pm$2.88 \\
    Synchronous &        35.15$\pm$3.74 &     38.83$\pm$6.24 \\
    \hline
          Total &         58.3$\pm$1.52 &     58.82$\pm$0.98 \\
\bottomrule
\end{tabular}
\caption{A comparison of F1 scores for the multi-sense method using cross-entropy and focal loss. RoBERTa$_{\text{large}}$ is used for the results. ``Total'' here refers to the average scores for all labels.}
\end{table}

Table 9 to Table 11 are the results for three methods in Macro-F1 scores, Precision, Recall and Hamming loss. 

In terms of Hamming loss, method 2 is the best, with a lowest Hamming loss of 5.70\%. For all methods, the label ``Purpose'' consistently shows the highest F1 scores, with precision and recall rates exceeding 92\%, indicating exceptional accuracy and consistency in identifying purpose relations. This uniformity suggests that all three methods are particularly adept at recognizing purpose-related discourse. For Method 1, most labels have a standard deviation in F1 score, recall, and precision under 10. 

However, some labels like ``Manner'' and ``Substitution'' show higher variability, indicating inconsistency in model performance for these relations. With regard to Method 2, ``Manner'' with a relatively high standard deviation in both precision and recall suggests variability in model performance across different folds. Method 3 shows lower precision of 15.06 ± 6.09, contrasting with a much higher recall of 55.07 ± 12.14. This stark difference implies challenges in accurately pinpointing ``Manner'' discourse relations. The ``Equivalence'' label also poses a challenge across methods, with notably poor performance in Method 1, reflecting difficulties in reliably identifying equivalence relations. Furthermore, labels like ``Purpose'', ``Condition'', and ``Asynchronous'' demonstrate lower standard deviations, indicating a more stable and consistent performance in these areas.

\section{Performance Variation Across Different Folds for Method 2}
Table 9 to Table 11 indicates a high degree of variability across folds in Precision, Recall or F1 scores. We can have a closer examination of individual labels for the Table 10. For instance, ``Manner'' exhibits high Recall (67.77 ± 9.88) but lower Precision (53.36 ± 15.81) and F1 score (58.60 ± 13.47), suggesting it identifies ``Manner'' instances well but has over-generalized Manner.``Synchronous'' consistently scores low across all metrics, indicating difficulty in identifying synchronous relationships. Labels like ``Equivalence'', ``Condition'', ``Substitutions'', and ``Manner'' have substantial standard deviations in Precision, with ±25.37, ±9.78 , ±10.45 and ±15.81 respectively, highlighting varying model performance across different test sets.

We thoroughly reviewed data and predictions for certain labels in each fold of our model's performance evaluation. For instance, in Fold 1, ``Equivalence'' achieved a perfect Precision of 100\% as the model correctly identified two true ``Equivalence'' instances. However, in other folds, Precision for ``Equivalence'' ranges from 10\% to 30\%, indicating significant variability in predictive accuracy.

\begin{figure*}
\centering
\includegraphics[width=12cm]{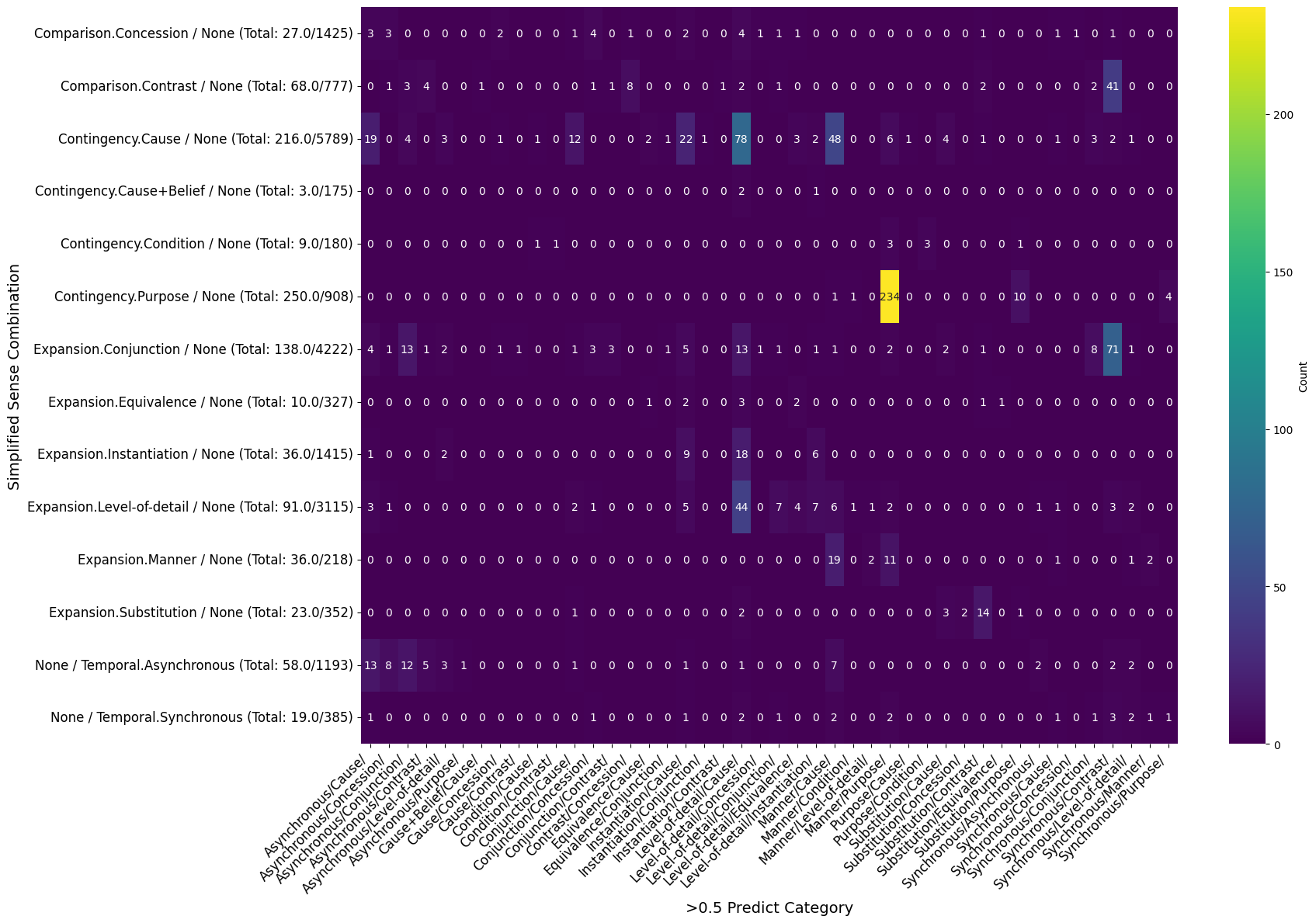}
\caption{Heatmap of overpredicted single label instances. This figure displays the distribution of instances where single-label is annotated but are given two labels by the model.}
\end{figure*}

\section{When Two labels Are Given to the Single-label Examples}
Figure 3 illustrates predictions of two labels for examples manually labeled with a single label. Vertical
axes correspond to gold labels, while horizontal
ones correspond to predicted labels. Numbers beside the vertical axis indicate the number of examples with that label and how many were predicted to have additional labels.

The figure reveals that the majority of labels predicted as two for single-labeled instances are relatively low for most labels. However,about 1/3 examples which only have the label "Purpose" are given the label ``Purpose'' and the label ``Manner'' simultaneously by the model. In addition, among the 218 examples whose label is ``Manner'', the model give the additional label ``Cause'' or ``Purpose'' for 30 examples. Besides, when the model give additional labels to those examples which are only annotated with ``Cause'', the labels are often under the Expansion category. 

These observations echo our earlier analysis, suggesting that distinguishing between single and dual labels poses challenges for models, particularly concerning ``Purpose''\&``Manner' and ``Purpose''. Additionally, models occasionally predict both elaborative and argumentative relations simultaneously although only one relation (elaborative or argumentative) is annotated, underscoring the potential value of incorporating multi-label annotation into the corpus.

\section{More Results on Using Focal Loss}
Table 12 demonstrates the results of cross-entropy and focal loss methods using RoBERTa$_{\text{b}}$. The results also show that focal loss could be better than cross-entropy for this task due to the data imbalance. 

Upon careful examination of Table 5 alongside this table, it becomes evident that focal loss exhibits advantages, particularly for labels with less annotated data such as ``Condition'', ``Equivalence'' and ``Synchronous''. Nevertheless, it's essential to acknowledge that while focal loss demonstrates usefulness for certain labels, its cannot improve performance across all labels . Consequently, future work can explore alternative methodologies to further enhance performance.

\end{document}